\documentclass[10pt, conference, compsocconf]{IEEEtran}
\ifCLASSINFOpdf
\else
\fi
\usepackage{cite}
\usepackage{amsmath,amssymb,amsfonts}
\usepackage{algorithmic}
\usepackage{graphicx}
\usepackage{textcomp}
\usepackage{xcolor}
\usepackage{subfigure}
\usepackage{array}
\usepackage{url}
\usepackage{multirow}
\usepackage{helvet}
\usepackage{courier}
\usepackage{amsmath}
\usepackage{amssymb}
\usepackage{booktabs}

\usepackage{subfigure}
\usepackage{algorithm}

\usepackage{multirow}
\usepackage{makecell} 
\usepackage{diagbox} 
\usepackage{cases}
\usepackage{threeparttable}
\usepackage{color}
\usepackage{caption}

\newtheorem{myDef}{Definition}

\begin{document}
%
\title{TTDM: A Travel Time Difference Model for Next Location Prediction}


\author{\IEEEauthorblockN{Qingjie Liu}
\IEEEauthorblockA{School of Software\\
Shandong University\\}
\and
\IEEEauthorblockN{Yixuan Zuo}
\IEEEauthorblockA{School of Computer Science and Technology\\
Shandong Jianzhu University\\}
\and
\IEEEauthorblockN{Xiaohui Yu}
\IEEEauthorblockA{School of Information Technology\\
York University\\}
\and
\IEEEauthorblockN{Meng Chen*}
\IEEEauthorblockA{School of Software\\
Shandong University\\
}
}


%


\maketitle

\begin{abstract}
Next location prediction is of great importance for many location-based applications and provides essential intelligence to business and governments. In existing studies, a common approach to next location prediction is to learn the sequential transitions with massive historical trajectories based on conditional probability. Unfortunately, due to the time and space complexity, these methods (e.g., Markov models) only use the just passed locations to predict next locations, without considering all the passed locations in the trajectory. In this paper, we seek to enhance the prediction performance by considering the travel time from all the passed locations in the query trajectory to a candidate next location. In particular, we propose a novel method, called Travel Time Difference Model (TTDM), which exploits the difference between the shortest travel time and the actual travel time to predict next locations. Further, we integrate the TTDM with a Markov model via a linear interpolation to yield a joint model, which computes the probability of reaching each possible next location and returns the top-rankings as results. We have conducted extensive experiments on two real datasets: the vehicle passage record (VPR) data and the taxi trajectory data. The experimental results demonstrate significant improvements in prediction accuracy over existing solutions. For example, compared with the Markov model, the top-1 accuracy improves by 40\% on the VPR data and by 15.6\% on the Taxi data.

\end{abstract}


%
\IEEEpeerreviewmaketitle

\section{Introduction}

With the increasing prevalence of electronic dispatch systems and video capturing equipments, it is possible to collect a deluge of traffic trajectory data. For example, with the widespread adoption of electronic dispatch systems, these mobile data terminals installed in each taxi could provide information on GPS (Global Positioning System)
localization and taximeter state. As another example, with the deployment of surveillance cameras on roads, vehicles are photographed when they pass the cameras and structured
vehicle passage records (VPRs) are subsequently extracted from the pictures using optical character recognition \cite{chen2016jcst,chennlpmm}. Such records contain at least three attributes: \textit{user}, \textit{location}, and \textit{time-stamp}. A consecutive sequence of these records from the same user constitute a trajectory. Learning from such trajectory data is an important task, and substantial progress in this domain can have a strong impact on many applications ranging from urban computing, traffic management, location-based recommendations and trajectory prediction \cite{he2018origin,zhao2018rest,chen2018pcnn,qiao2015traplan,yin2017spatial,bao2017planning,chen2018mpe,zhao2016predicting}.

\begin{figure*}[!tb]
\centering
\includegraphics[width=0.98\textwidth]{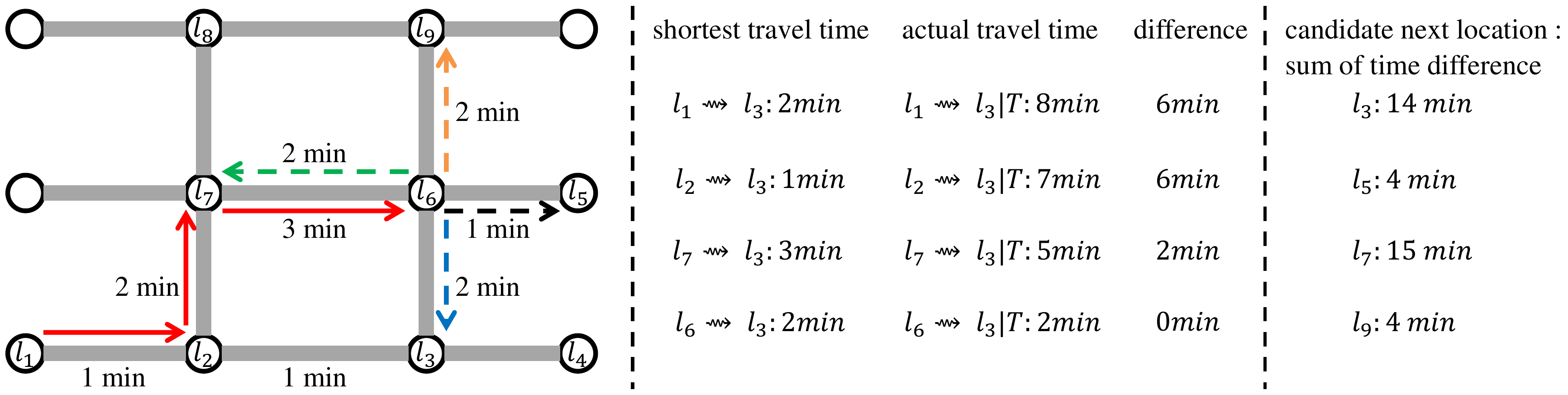}
\caption{An illustrative example of next location prediction. As shown in the left part, the user has visited location $l_1@9:05$ (short for $l_1$ at 9:05 a.m.), $l_2@9:06$, $l_7@9:08$, and $l_6@9:11$. What is his/her next visited location, $l_3$, $l_5$, $l_7$, or $l_9$? Given a query trajectory sequence $T: l_1 \rightarrow l_2 \rightarrow l_7 \rightarrow l_6$, taking a candidate next location $l_3$ as an example, we set 1 minute between $l_2$ and $l_3$, it means that the average travel time of $l_2 \rightarrow l_3$ and $l_3 \rightarrow l_2$ are both 1 minute. We compute the shortest travel time from all the passed locations in T to $l_3$ based on the average travel time from one location in T to its neighbors, and the actual travel time to $l_3$ along $T$, as shown in the central part. We list all the candidate next locations and compute the travel time difference, and predict next locations based on it, as shown in the right part.}
\label{intro}
\end{figure*}

One of the fundamental problem in trajectory mining is next location prediction. That is, given a query trajectory sequence, predicting the next location that a user will arrive at, as shown in Fig.~\ref{intro}. Next location prediction is of great value to both users and the owners of trajectory datasets. For example, if we know the successive locations that users intend to visit in advance, we could optimize marketing strategies accordingly and push promotions to those in the predicted area. Furthermore, such knowledge may also assist in forecasting traffic conditions and routing the drivers so as to alleviate traffic jams.

To predict next locations, one popular method is to learn the sequential transitions based on a Markov model \cite{chen2015mining} or frequent patterns \cite{monreale2009wherenext}.
For example, Chen et al. \cite{chen2015mining} used a Markov model to mine both individual mobility patterns and collective mobility patterns to predict next locations. Monreale et al. \cite{monreale2009wherenext} built a T-pattern tree with all the trajectories to make future location predictions. However, due to the space and time complexity, these methods only consider the just passed locations in mining human mobility patterns, without considering the whole locations in the trajectory, which may hinder the prediction performance. For example, as shown in Fig.~\ref{intro}, given a trajectory sequence $l_1 \rightarrow l_2 \rightarrow l_7 \rightarrow l_6$, if we predict next locations based on the current location $l_6$, we will know that the user has equal probability to arrive at $l_3$, $l_5$, $l_7$, or $l_9$ next. If we pay attention on the whole trajectory sequence, we will infer that the user will be more likely to visit $l_5$ or $l_9$ next, as the user has more reasonable routes to arrive at $l_3$ and $l_7$ from the origin location $l_1$.

As we know, users' moving purposes directly point to the destination and they usually choose an optimal path as a moving route to link two locations. Back to Fig.~\ref{intro}, the user could arrive at location $l_3$ from $l_1$ in two minutes via a path of $l_1 \rightarrow l_2 \rightarrow l_3$, and it has a low probability that the user chooses a longer path $l_1 \rightarrow l_2 \rightarrow l_7 \rightarrow l_6 \rightarrow l_3$ within eight minutes. Therefore, we propose to model the travel time from all the passed locations in the query trajectory to a candidate next location to enhance the prediction performance.

In this paper, we propose a Travel Time Difference Model (TTDM for short) to predict next locations. TTDM first scans the historical trajectory data, and builds the weighted location transfer graph with the average travel time of each location transition. In the weighted location transfer graph, the location is referred as to the node, and the location transition is referred to as the edge. Next, TTDM calculates the shortest travel time between any two locations via Yen's algorithm \cite{yen1971finding}. Further, given a query trajectory sequence, TTDM computes the actual travel time between the passed locations and a candidate next location, and fetches the corresponding shortest travel time calculated in advance. Finally, TTDM predicts the next locations in terms of the difference between the actual travel time and the shortest travel time. To summarize, one distinct feature of TTDM is that it could capture the effect of long distance prefix locations. An illustrative example of our TTDM on next location prediction is shown in Fig.~\ref{intro}.

Further, we know that the next location is also affected by the just passed locations, and the Markov-based methods have performed well in the task of next location prediction \cite{chen2015mining}. The Markov models focus on the \textbf{local} sequential transitions, and our TTDM mines the \textbf{global} information in the trajectory, and the two models
contribute differently into the probability of visiting a candidate next location. Therefore, we use a linear interpolation to balance the two models, and yield a joint model. When making prediction, we compute the probability of reaching each possible next location, and return the top ranking results as outputs.

We present experimental results on two real datasets. One consists of the vehicle passage records over a period of two months and another contains the taxi trajectories in one year. The experimental results confirm the superiority of the proposed models over existing methods.

The contributions of this paper can be summarized as follows.

\begin{itemize}
\item We present a Travel Time Difference Model, which considers the shortest travel time and the actual travel time from the passed locations in the query trajectory sequence to a candidate  next location and leverages the travel time difference to predict next locations. To the best of our knowledge, it is the first work that uses the global information in the query trajectory and the travel time to enhance the prediction accuracy.
\item We integrate our TTDM with a Markov model via a linear interpolation to yield a joint model. The joint model considers both the local sequential transitions and the global travel time information, and obtains the best performance. Our model could also integrate with other location prediction models, e.g., the embedding based model \cite{chen2018mpe}, the recurrent neural network based model \cite{liu2016predicting}.
\item We conduct extensive experiments with real-world traffic data to investigate the effectiveness of the proposed models, showing remarkable improvement as compared with baselines in predicting next locations. As a side contribution, we have released the datasets and codes to facilitate the community research \footnote{https://github.com/tracyitbird/TTDM}.
\end{itemize}

The rest of the paper is organized as follows. We review the related work in Section.~\ref{relatedwork}, and give the preliminaries of our work in Section.~\ref{pre}. In Section.~\ref{ttdmsec}, we introduce the Travel Time Difference Model for next location prediction. In Section.~\ref{jointmodel}, we integrate our TTDM with a Markov model to generate a  joint model. We present the experimental results and the performance analysis in Section.~\ref{experiment}, and conclude this paper in Section.~\ref{conclusion}.

\section{Related Work}
\label{relatedwork}

Trajectory data mining has become a hot research topic recently, and Zheng \cite{zheng2015trajectory} has conducted a systematic survey on this topic. In this paper, we aim at predicting the next locations, and mainly discuss the recent progress in this field.

\subsection{Prediction with Markov models}
Most of conventional methods adopt Markov models to mine the human mobility patterns to make next location prediction. For example, Xue et al. \cite{xue2009traffic} used taxi traces to construct a Probabilistic Suffix Tree and predicted short-term routes with variable-order Markov Models. Chen et al. \cite{chen2015mining} proposed to mine both individual and collective movement patterns with an integrated variable-order Markov model to predict the successive locations. Ye et al. \cite{ye2013s} proposed a framework which uses a mixed hidden Markov model to predict the category of user activity and then predict the most likely location given the estimated category distribution.

\subsection{Prediction with deep learning models}
Recently, there also exist some methods using the recurrent neural networks to model the sequential patterns in trajectory data for the location prediction. For instance, Liu et al. \cite{liu2016predicting} proposed a method called Spatial Temporal Recurrent Neural Networks (ST-RNN), which models the local temporal and spatial contexts in each layer for mining mobility patterns. Yang et al. \cite{yang2017neural} mined both the social networks and mobile trajectories in a neural network, in which they employed RNN to capture the sequential relatedness in mobile trajectories. Kong and Wu \cite{kong2018hst} proposed a hierarchical spatial-temporal LSTM model, leveraging the historical visit information and spatial-temporal factors for the location prediction.

In addition, some works \cite{zhou2016general,feng2015personalized, zhao2017geo, chen2018mpe} adopt embedding methods to make successive location prediction. For example, Zhou et al. \cite{zhou2016general} proposed a Multi-Context Trajectory Embedding Model (MC-TEM) based on the framework of \textit{word2vec} \cite{le2014distributed}, and considered various useful contextual features, including user-level, trajectory-level, location-level and temporal contexts to make point-of-interest (POI) prediction. Feng et al. \cite{feng2015personalized} proposed a personalized ranking metric embedding method (PRME) to make POI prediction, which first embeds each POI into a sequential transition space and then projects each POI and user into a user preference space. Zhao et al. \cite{zhao2017geo} adopt the framework of \textit{word2vec} by treating each user as a ``document'', check-ins in a day as a ``sentence'', and each POI as a ``word'', and proposed a Geo-Temporal sequential embedding rank (Geo-Teaser) model for POI recommendation. Chen et al. \cite{chen2018mpe} focused on the traffic trajectory data and proposed a Mobility Pattern Embedding (MPE) method to embed the time slots, current locations and next locations together as points in a latent space, and predicted the next locations based on the embedding vectors.

In this paper, we integrate our model with a Markov model to generate a joint model to predict next locations. Integrating our model with these recent deep learning models is our future work.

\subsection{Prediction with external information}
Pushing further from the historical trajectories, there are studies \cite{Zhang2015NextMe, zhou2013semi, pan2012utilizing } that improve prediction performance with external information (e.g., semantic features, driving speed and direction). Zhang et al.\cite{Zhang2015NextMe} extracted the underlying correlation between human mobility patterns and cellular call patterns and made location prediction based on it from temporal and spatial perspectives. Given a target trajectory, Zhou et al. \cite{zhou2013semi} extracted a small set of reference trajectories and trained a local model for prediction. Pan et al. \cite{pan2012utilizing} incorporated the historical traffic data with the real-time event, and proposed H-ARIMA+ to predict traffic in the presence of incidents. In this paper, we consider the travel time of each road segment, and use the difference between the shortest travel time and the actual travel time to predict next locations. To the best of our knowledge, this is the first paper that leverages travel time to enhance the prediction performance.

\section{Preliminaries}
\label{pre}
In this section, we first introduce some concepts which are required for the subsequent discussion, and then give an overview of the problem addressed in this paper, and finally list the major notations in Tab.~\ref{tab:notation}.

\begin{myDef}[Location] A  user may visit a set of {\em locations}, where each location $l$ refers to a point or a region where the position of the user is recorded.
\end{myDef}

\begin{myDef}[Trajectory] The {\em trajectory} $T$ is defined as a time-ordered sequence of locations: $\langle (l_{1},t_1), (l_{2},t_2), \ldots, \\(l_{n},t_n) \rangle$, where $t_i$ is the time-stamp that a user arrives at location $l_i$.
\end{myDef}

\begin{myDef}[Moving Segment] A {\em moving segment} $s_i$ is represented as a directed segment that contains two consecutive locations $l_i$ and $l_{i+1}$ in a trajectory, i.e., $s_i = l_i\rightarrow l_{i+1}.$
\end{myDef}

\begin{myDef}[Candidate Next Locations] Given a location $l_{i}$, we define location $l_{j}$ as a {\em candidate next location} of $l_{i}$ if a user can arrive at $l_{j}$ from $l_{i}$ without visiting another location first.
\end{myDef}

The set of candidate next locations can be obtained either by prior knowledge (e.g., locations of the surveillance cameras combined with the road network graph), or by the induction from historical trajectories of moving objects.

Given a query trajectory sequence $T=\langle (l_{1},t_1), (l_{2},t_2), \\ \ldots, (l_{n},t_n) \rangle$, the next location prediction problem is to predict the location that a user will arrive at next. That is, given $T$, to predict the next location $l_{n+1}$.

\begin{table}
\small
\centering
\caption{Notations and descriptions.}
\renewcommand{\arraystretch}{1.5}
\begin{tabular}{ l| l}
\Xhline{1pt}
Notations & Descriptions \\
\hline
$T$ & trajectory  \\
$s$ &  moving segment  \\
$l, t$ &  location, time  \\
$\overline{t}(s_i)$ &  the average travel time of moving segment $s_i$ \\
$\Delta t_{s}(l_i \rightsquigarrow l_j)$ & the shortest travel time from $l_i$ to $l_{j}$\\
$\Delta t_{a}(l_i \rightsquigarrow l_j|T)$ & the actual travel time from $l_i$ to $l_{j}$ along $T$\\
\Xhline{1pt}
\end{tabular}
\label{tab:notation}
\end{table}

\section{Travel Time Difference Model}
\label{ttdmsec}
Generally speaking, users' moving purposes directly point to the destination. Under the given conditions of origin and destination, by incorporating some factors (e.g., travel distance, travel time, energy consumption), users usually choose an optimal path as a moving route to link these two locations. If we examine the historical trajectories, we will find that users like to choose a route which costs relatively less time from the passed locations to the targeted next location. For example, as shown in Fig.~\ref{intro}, users are more likely to arrive at location $l_3$ from $l_2$ directly within 1 minute, instead of taking a path of $l_2 \rightarrow l_7 \rightarrow l_6 \rightarrow l_3$ which will cost 7 minutes. Further, given a trajectory sequence $(l_1, l_2, l_7, l_6)$, $l_3$ has a low probability to be the next location. Therefore, we could leverage the travel time between the passed locations and the candidate next locations to help improve the prediction accuracy.

\begin{figure}[!t]
\centering
\includegraphics[width=0.45\textwidth]{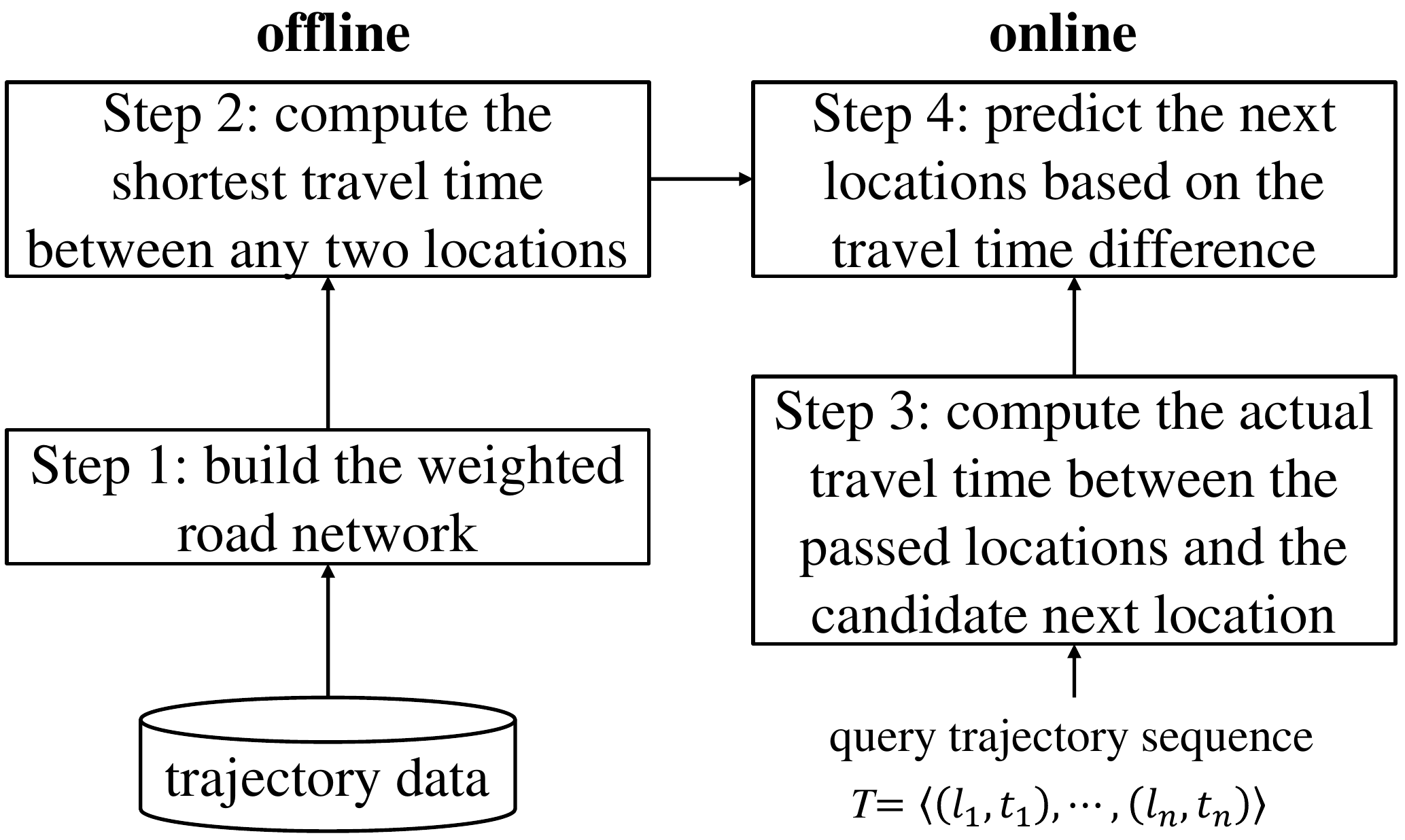}
\caption{The framework of our travel time difference model.}
\label{framework}
\end{figure}

To make next location prediction, we propose a \textbf{Travel Time Difference Model}, named TTDM, whose framework is shown in Fig.~\ref{framework}. Based on the historical trajectory data, we first build the weighted location transfer graph with the average travel time of moving segments, and then compute the shortest travel time between any two locations via Yen's algorithm \cite{yen1971finding}. These work could be completed offline in advance. Further, given a query trajectory sequence, we compute the actual travel time between the passed locations and a candidate next location. At the same time, we fetch the shortest travel time between the passed locations and the candidate next location calculated in step 2. Finally, we predict the next locations in terms of the difference between the actual travel time and the shortest travel time. In the following subsection, we will introduce the four components in detail.

\textbf{1. Build the weighted location transfer graph based on the average travel time of moving segments.} Given a moving segment $s_i$, we could compute the average travel time $\overline{t}(s_i)$ of vehicles for that segment based on the historical trajectory data. As the travel time $\overline{t}(s_i)$ of a segment varies with time, we could compute the travel time for each time slot, e.g., $\overline{t}^{k}(s_i)$ for the $k$th slot. Based on the segments and locations, we regard the road network as a graph, where the location is referred to as the node, the segment is the edge, and the average travel time $\overline{t}(s_i)$ is the weight of the edge. Supposed we set the size of time slot at 30 minutes, we will have 48 slots in a day, and get 48 snapshots for weighted location transfer graphs.

\textbf{2. Compute the shortest travel time between two locations.} Based on the $k$th weighted location transfer graph, we calculate the shortest travel time $\Delta t_{s}^{k}(l_i \rightsquigarrow l_j)$ between two locations $l_i$ and $l_j$ according to Yen's shortest path algorithm \cite{yen1971finding}. Yen's algorithm computes single-source $K$-shortest loopless paths for a graph with non-negative edge cost. Given an origin location $l_i$, we first choose the $k$th weighted road network based on the time-stamp that a user arrives at $l_i$, and then compute the shortest travel time $\Delta t_{s}^{k}(l_i \rightsquigarrow l_j)$ to the target location $l_j$. Note that, Yen's algorithm is time-consuming, and we compute the shortest travel time between any two possible locations with the weighted road networks in advance, and save the values in an efficient data structure (e.g., key-value pairs) to support real-time applications.

\begin{figure}[!t]
\centering
\includegraphics[width=0.45\textwidth]{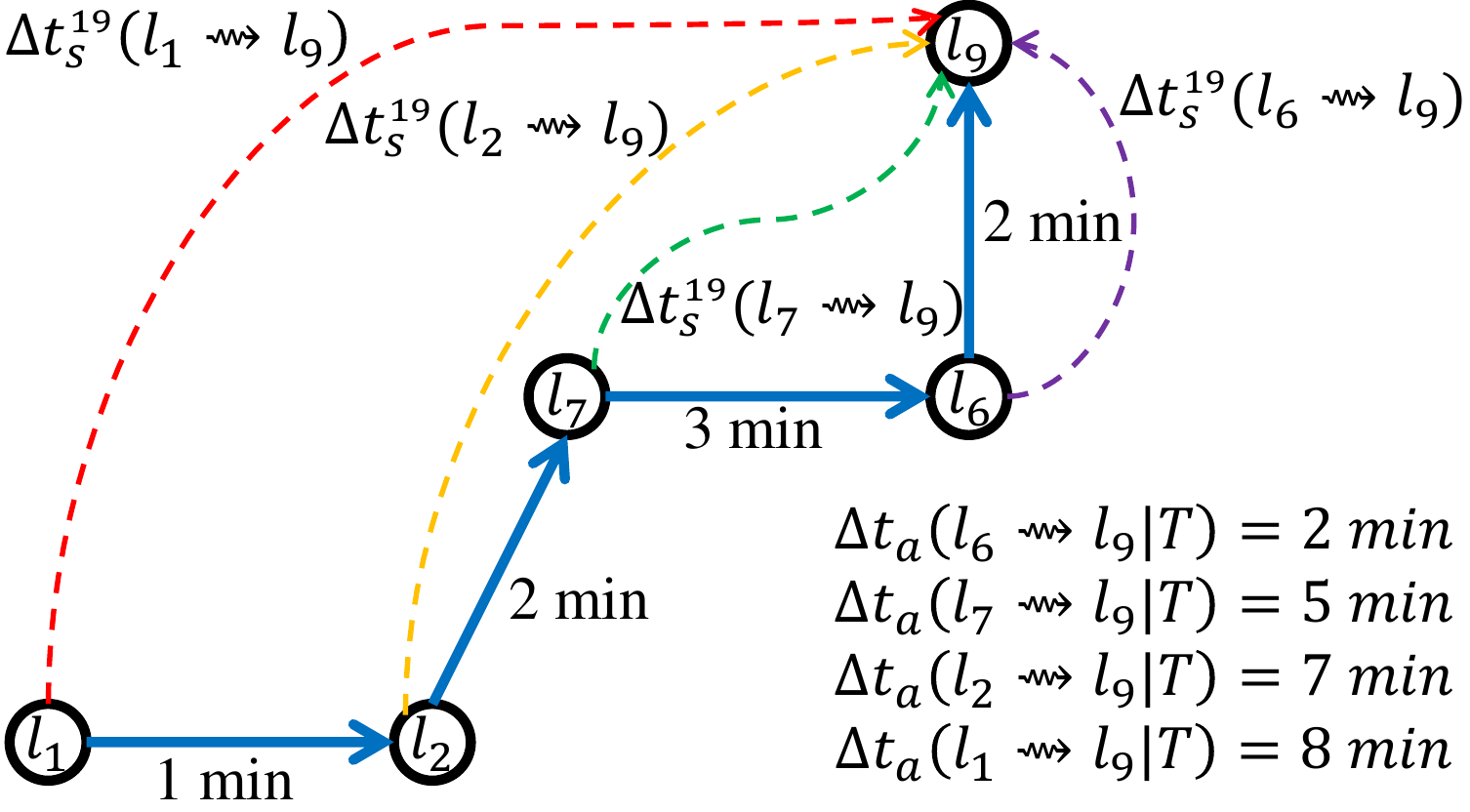}
\caption{An illustrative example of a user's trajectory. The user has visited location $l_1@9:05$, $l_2@9:06$, $l_7@9:08$, $l_6@9:11$, and $l_9@9:13$, and the actual travel time between any two locations could be computed accordingly, e.g., $\Delta t_{a}(l_1 \rightsquigarrow l_9|T) = (9:13-9:05)=8 \ min$. According to Yen's shortest path algorithm \cite{yen1971finding}, we could compute the shortest travel time between any two locations in the nineteenth time slot (here we choose 30 minutes as the size of a time slot, and define 0:00-0:30 as the first slot.), e.g., $\Delta t_{s}^{19}(l_1 \rightsquigarrow l_9)$. Note that, the shortest travel time $\Delta t_{s}^{k}(l_i \rightsquigarrow l_j)$ between two locations $l_i$ and $l_j$ in the $k$th slot may be equal to the actual travel time $\Delta t_{a}(l_i \rightsquigarrow l_j)$.}
\label{traExample}
\end{figure}

\textbf{3. Compute the actual travel time between the passed locations and the candidate next location.}
Given a query trajectory sequence $T=\langle (l_{1},t_{1}), \ldots, (l_{i},t_{i}), \ldots, (l_{n},t_{n}) \rangle$, we could compute the actual travel time $\Delta t_{a}(l_i \rightsquigarrow l_{n+1}|T)$ from the passed location $l_i$ to a candidate next location $l_{n+1}$ following the trajectory $T$, i.e.,
\begin{equation}
\Delta t_{a}(l_i \rightsquigarrow l_{n+1}|T) = t_{n}-t_i + \overline{t}(l_n \rightarrow l_{n+1}).
\end{equation}
As we do not know the actual travel time between the current location $l_n$ and the candidate next location $l_{n+1}$ indeed, we use the average travel time $\overline{t}(l_n \rightarrow l_{n+1})$ instead. An example of the shortest travel time and the actual travel time is illustrated in Fig.~\ref{traExample}.

\textbf{4. Predict the next locations based on travel time difference.}
As mentioned earlier, in purpose-driven moving scenario, users always choose the shortest route toward its actual destination, and the travel time from the passed location to the candidate next location along the route is relatively the least. Hence, given a trajectory $T=\langle (l_{1},t_{1}), \ldots, (l_{n},t_{n}) \rangle$ and its candidate next location $l_{n+1}$, if the shortest travel time $t_{s}^{k}(l_i \rightsquigarrow l_{n+1})$ from the passed location $l_i$ to the candidate next location $l_{n+1}$ is less than the actual time $t_{a}(l_i \rightsquigarrow l_{n+1}|T)$ evidently, $l_{n+1}$ has the less likely to be the expected next location. In order to quantify and evaluate the strength of mobility approximation for each candidate next location, we leverage the aforementioned definitions to calculate and compare the actual travel time and the shortest travel time location by location in the trajectory.

For each passed location in the trajectory $T=\langle (l_{1},t_{1}), \ldots, (l_{n},t_{n}) \rangle$, we compute the corresponding shortest travel time with regard to a candidate next location $l_{n+1}$. Thus the summation of the travel time from all the passed locations to the candidate $l_{n+1}$ can be regarded as the corresponding strength of mobility approximation for this specific next location, defined as
\begin{equation}
\Delta t_{s}^{sum} = \sum_{i=1}^{n} \Delta t^{k}_{s}(l_i \rightsquigarrow l_{n+1}).
\end{equation}
We need to determine the $k$th weighted location transfer graph based on the time-stamp.

Similarly, the summation of the travel time from all the passed locations to the candidate $l_{n+1}$ along the trajectory $T$ is defined as
\begin{equation}
\Delta t_{a}^{sum} = \sum_{i=1}^{n} \Delta t_{a}(l_i \rightsquigarrow l_{n+1}|T).
\end{equation}

Finally, given a trajectory $T=\langle (l_{1},t_{1}), \ldots, (l_{n},t_{n}) \rangle$, the probability of visiting a candidate next location $l_{n+1}$ is
\begin{equation}
\label{ttdm}
p\left( l_{n+1}|T \right) = \dfrac{1}{Z(T)} f\left(\dfrac{\Delta t_{a}^{sum} - \Delta t_{s}^{sum}}{|T|}\right),
\end{equation}
where $|T|$ represents the number of locations in the trajectory $T$, and $Z(T)$ is the normalization term and it is computed by
\begin{equation}
Z(T)= \sum_{i=1}^{m}f^p\left(\dfrac{\Delta t_{a}^{sum} - \Delta t_{s}^{sum}}{|T|}\right),
\end{equation}
where $m$ represents the number of candidate next locations and $f^p\left(\dfrac{\Delta t_{a}^{sum} - \Delta t_{s}^{sum}}{|T|}\right)$ represents the function $f$
acts on the $p$th candidate next location. For the function $f$, we could choose different inverse function, e.g.,
\begin{equation}
\label{reverse}
\begin{aligned}
f(x) &= \exp(-x), \\
f(x) &= \dfrac{1}{x}.
\end{aligned}
\end{equation}

\section{Integrate TTDM with a Markov Model}
\label{jointmodel}
As we know, the next location is mainly affected by the just passed locations, and many methods adopt a Markov model to make successive location prediction \cite{qiao2018hybrid, chen2015mining}. The Markov model regards the trajectory of a user as a discrete stochastic process. Specifically, a state in the Markov model corresponds to a location, and a state transition corresponds to moving from one location to the other. In this section, we first introduce the Markov model briefly, and then integrate it with our TTDM to predict next locations.

Given a trajectory $T=\langle (l_{1},t_{1}), \ldots, (l_{n},t_{n}) \rangle$, let $p\left( l_{n+1}|T\right)$ be the probability that the user will arrive at location $l_{n+1}$ next,
\begin{equation}
p\left(l_{n+1}|T \right)  =  p\left(l_{n+1}|l_{1},l_{2}, \dots, l_{n} \right).
\end{equation}
Essentially, this approach for each candidate next location $l_{n+1}$ computes its probability of next visit, and selects the one that has the highest possibility.

According to the memorylessness of a Markov chain, the probability of arriving at $l_{n+1}$ actually only depends on the just passed locations, instead of all of them in the trajectory.
Therefore, the probability that the user will arrive at location $l_{n+1}$ next can be given by
\begin{equation}
p\left(l_{n+1}|T \right) =  p\left(l_{n+1} | l_{n} \right).
\end{equation}

In order to use the Markov model, we learn $l_{n+1}$ for each prefix location $l_n$, by estimating the conditional probability $p\left(l_{n+1} | l_{n} \right)$. The most commonly used method for estimating this value is to use the maximum likelihood principle, and the conditional probability $p\left(l_{n+1} | l_{n} \right)$ therefore can be computed by
\begin{equation}
\label{mm}
p\left(l_{n+1} | l_{n} \right) = \dfrac{\sharp (l_n,l_{n+1})}{\sharp (l_{n})},
\end{equation}
where $\sharp (l_n)$ is the number of times that location $l_n$ occurs in the training set, and $\sharp (l_{n},l_{n+1})$ is the number of times that location $l_{n+1}$ occurs immediately after $l_n$.

Given a trajectory $T=\langle (l_{1},t_{1}), \ldots, (l_{n},t_{n}) \rangle$, the TTDM models the travel time of all the passed locations to a candidate next location, and it considers the long distance prefix locations; the Markov model learns the sequential transition for a candidate next location from the prefix location. As two components contribute differently into the probability of visiting a candidate next location, we use a linear interpolation to weight the two models based on Equation.~(\ref{ttdm}) and Equation.~(\ref{mm}),
\begin{equation}
\small
p\left(l_{n+1}|T \right) =  \lambda \dfrac{\sharp (l_n,l_{n+1})}{\sharp (l_{n})} + \dfrac{1-\lambda}{Z(T)}f\left(\dfrac{\Delta t_{a}^{sum} - \Delta t_{s}^{sum}}{|T|}\right),
\end{equation}
where $\lambda \in [0,1]$ controls the weight of different kinds of models.

Note that, except the Markov model, many other location prediction methods (e.g., Bayes models, embedding models) could also be integrated with our TTDM \cite{chen2018mpe, chang2018content, liu2016predicting, zhao2017time}.

\section{Experiment}
\label{experiment}
To evaluate the performance of our models, we have conducted extensive experiments on two real  datasets. In this section, we first introduce the data and settings in our empirical study, followed by the results of our methods and baselines. Finally, we show the performance evaluated with the parameters.
\subsection{Datasets and Settings}
In the experiments, we use two datasets: the vehicle passage record (VPR) data and the taxi trajectory data. The datasets are released to facilitate the community research \footnote{https://github.com/tracyitbird/TTDM}.

\begin{table}[!t]
\centering
\caption{Data statistics}
\renewcommand{\arraystretch}{1.5}
\begin{tabular}{l|c|c}
\Xhline{1pt}
&   VRP Data &   Taxi Data  \\
\hline
$\sharp$users & 15,321 & 439\\
$\sharp$locations & 360 & 902\\
$\sharp$trajectories & 287,605 & 633,063\\
average length of each trajectory & 6.2 & 15.7\\
\Xhline{1pt}
\end{tabular}
\label{tab:datastatistic}
\end{table}

\begin{figure}[!ht]
\centering
\includegraphics[width=0.35\textwidth]{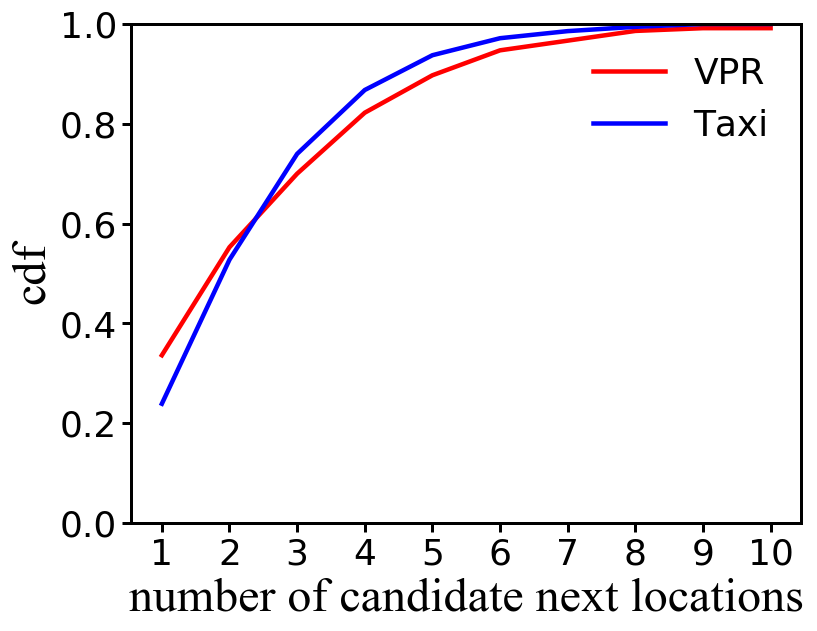}
\caption{Characteristics of the trajectory data.}
\label{cdf}
\end{figure}

\begin{table*}[!t]
\centering
\renewcommand{\arraystretch}{1.5}
\caption{Results of methods on the VPR data.}
\begin{threeparttable}
\begin{tabular}{ p{0.08\textwidth}| p{0.08\textwidth}<{\centering}| p{0.08\textwidth}<{\centering}| p{0.1\textwidth}<{\centering}| p{0.08\textwidth}<{\centering}| p{0.08\textwidth}<{\centering}| p{0.1\textwidth}<{\centering} }
\Xhline{1pt}
\multirow{2}*{} & \multicolumn{3}{c|}{Acc} & \multicolumn{3}{c}{Ap}  \\
\cline{2-7}
\multirow{2}*{} & MM & TTDM  & TTDM+MM  & MM & TTDM  & TTDM+MM \\
\cline{2-7}
\hline
top-1  & 0.417 & 0.527 & \textbf{0.584 } & 0.417 & 0.527 & \textbf{0.584}    \\

top-2  & 0.713 & 0.738 & \textbf{0.793}   & 0.565  & 0.632 & \textbf{0.688}  \\

top-3  & 0.830 & 0.850 & \textbf{0.889 } & 0.604 & 0.670 & \textbf{0.720}    \\

top-4  & 0.901 & 0.903 & \textbf{0.934}  & 0.622 & 0.683 & \textbf{0.732}    \\

top-5  & 0.948 & 0.957 & \textbf{0.965} & 0.631 & 0.694 & \textbf{0.738}     \\

\Xhline{1pt}
\end{tabular}
\begin{tablenotes}
\footnotesize
\item[1] The improvements over the baselines are statistically significant in terms of paired t-test \cite{hull1993using} with $p$ value $<$ 0.01.
\end{tablenotes}
\end{threeparttable}
\label{tab:baseline1}
\end{table*}

\begin{table*}[!ht]
\centering
\renewcommand{\arraystretch}{1.5}
\caption{Results of methods on the Taxi data.}
\begin{threeparttable}
\begin{tabular}{ p{0.08\textwidth}| p{0.08\textwidth}<{\centering}| p{0.08\textwidth}<{\centering}| p{0.1\textwidth}<{\centering}| p{0.08\textwidth}<{\centering}| p{0.08\textwidth}<{\centering}| p{0.1\textwidth}<{\centering}}
\Xhline{1pt}
\multirow{2}*{} & \multicolumn{3}{c|}{Acc} & \multicolumn{3}{c}{Ap}  \\
\cline{2-7}
\multirow{2}*{} & MM & TTDM  & TTDM+MM  & MM & TTDM  & TTDM+MM \\
\cline{2-7}
\hline
top-1  & 0.500 & 0.538 & \textbf{0.578 } & 0.500 & 0.539 & \textbf{0.578} \\

top-2  & 0.781 & 0.791 & \textbf{0.817}   & 0.640  & 0.665 & \textbf{0.698} \\

top-3  & 0.895 & 0.915 & \textbf{0.930 } & 0.678 & 0.706 & \textbf{0.735} \\

top-4  & 0.960 & 0.972 & \textbf{0.976}  & 0.695 & 0.721 & \textbf{0.747} \\

top-5  & 0.986 & 0.990 & \textbf{0.992} & 0.700 & 0.724 & \textbf{0.750} \\

\Xhline{1pt}
\end{tabular}
\begin{tablenotes}
\footnotesize
\item[1] The improvements over the baselines are statistically significant in terms of paired t-test \cite{hull1993using} with $p$ value $<$ 0.01.
\end{tablenotes}
\end{threeparttable}
\label{tab:baseline2}
\end{table*}

\textbf{VPR data}: We collect two months (09/2015 - 10/2015) of VPRs over the traffic surveillance system in a major metropolitan city. Each record is extracted
from the picture using optical character recognition (OCR), containing a vehicle ID, the location of the surveillance camera (location: $l$), and the time-stamp of the vehicle passing location $l$ (time: $t$). We take all the records and pre-process them to form trajectories. To make the model more robust, we include only those trajectories that contain at least three locations. The detailed information is shown in Tab.~\ref{tab:datastatistic}, where $\sharp$users means the number of unique users, and the others have the similar meaning. In this way, we obtain a total of 287,605 trajectories, and each trajectory contains 6.2 locations on average.

\textbf{Taxi data}: The taxi data is composed of all the complete trajectories of 442 taxis running in the city of Porto (Portugal) of 389 sq.km for a complete year (from 01/07/2013 to 30/06/2014). We discretize the region of interest into a grid with equal-sized cells, and assign a cell index for each GPS location. We pre-process the dataset to generate trajectories, and only keep those that contain at least three locations. As shown in Tab.~\ref{tab:datastatistic}, the taxi data contains 633,063 trajectories, in which each trajectory has 15.7 locations on average.

To understand the data better, we compute the cumulative distribution function (cdf) of the number of candidate next locations, as shown in Fig.~\ref{cdf}. It is clear that about 44.7\% of the locations have more than two candidate
next locations on the VPR data, and the average number of candidate next locations is about 2.75. On the Taxi data, about 47.2\% of the locations have more than two candidate locations, and the average number of candidate next locations is about 2.68.

For both datasets, we randomly split them into two collections in proportion of 8:2 as the training set and test set, and perform 10 runs (with the same data split) to report the average of the experimental results.  In the experiments, given a query trajectory sequence $T$, we predict the top-$r$ next locations. That is, selecting the top-$r$ locations $l_{n+1}$ that have the highest possibility $p(l_{n+1}|T)$. As mentioned in Equation.~(\ref{reverse}), we choose $f(x)=\dfrac{1}{x}$ as the reverse function in our experiments. As the trajectories in one hour is not enough, we set the size of time slot at 24 hours in TTDM, and build a weighted location transfer graph. If we have sufficient trajectories, we could build multiple weighted location transfer graphs according to the size of time slot. The default value for the weight $\lambda$ that balances TTDM and the Markov model is 0.3, and we will evaluate the effect of this parameter in the experiments.

To compare different methods, we use two well known metrics following \cite{chen2015mining}, namely, \textit{accuracy} and \textit{average precision} (denoted by \textit{acc} and \textit{ap} respectively), to evaluate the prediction performance. Accuracy is defined as the ratio of the number of trajectories for which the model is able to correctly predict to the total number of trajectories in the test set. That is,
\begin{equation}
acc = \frac{1}{\left| \tau' \right|}\sum {p(l)},
\end{equation}
where $|\tau'|$ is the number of trajectories in the testing set. Let $w$ denote the rank of the actual next location at each prediction, and $p(l)$ is 1 if $w \leq r$ (i.e., the actual next location can be found in top $r$ ranked positions) and 0 otherwise.

Average precision is defined as
\begin{equation}
 ap = \frac{1}{\left| \tau' \right|}\sum {\dfrac{p(l)}{w}},
\end{equation}
where $w$ is the rank of the actual next location at each prediction, and $p(l)$ is 1 if $w \leq r$. Average precision puts a larger weight to the top-ranked actual next location.

\begin{figure*}[!t]
\centering
\subfigure[Accuracy]{
\includegraphics[width=0.35\textwidth]{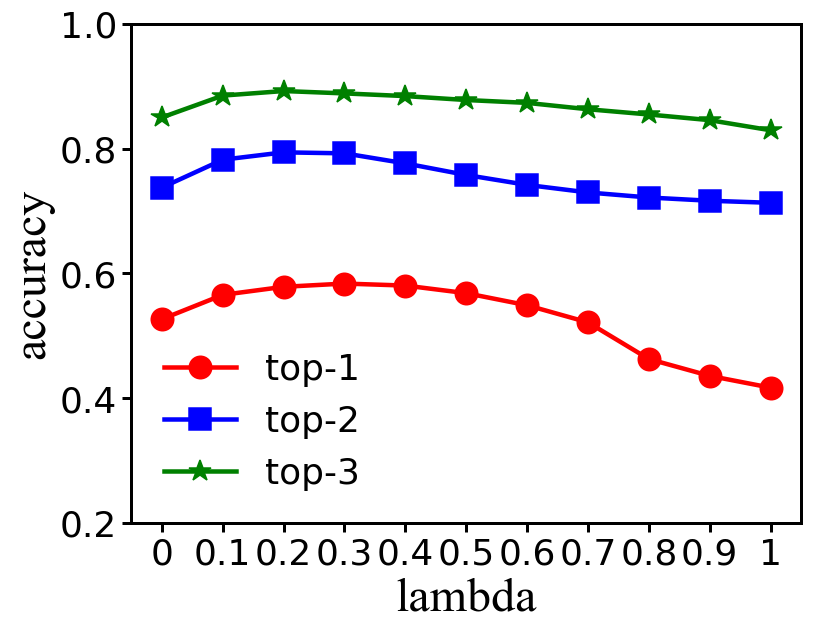}}
\hspace{0.1\textwidth}
\subfigure[Average Precision]{
\includegraphics[width=0.35\textwidth]{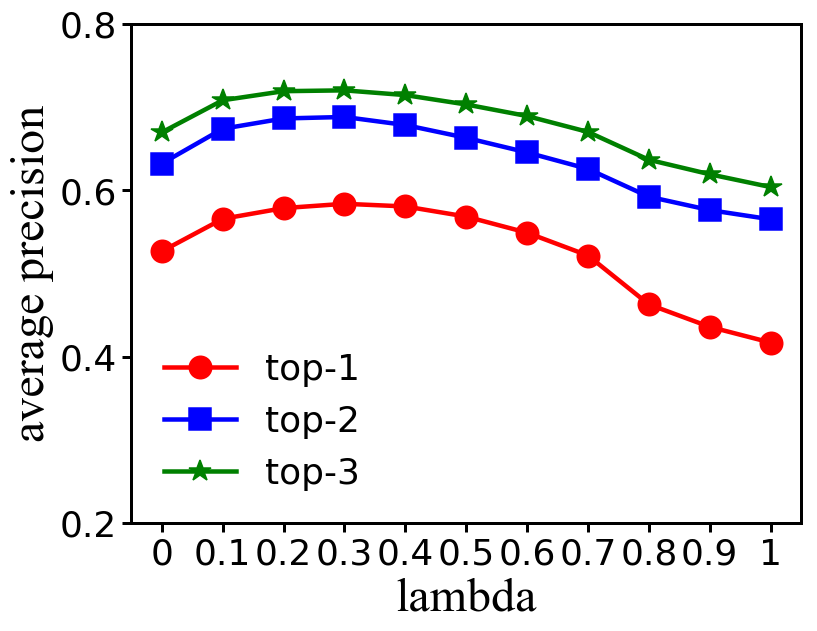}}
\caption{Performance of the joint model with different $\lambda$ on the VPR data. }
\label{fig:lambda1}
\end{figure*}

\begin{figure*}[!t]
\centering
\subfigure[Accuracy]{
\includegraphics[width=0.35\textwidth]{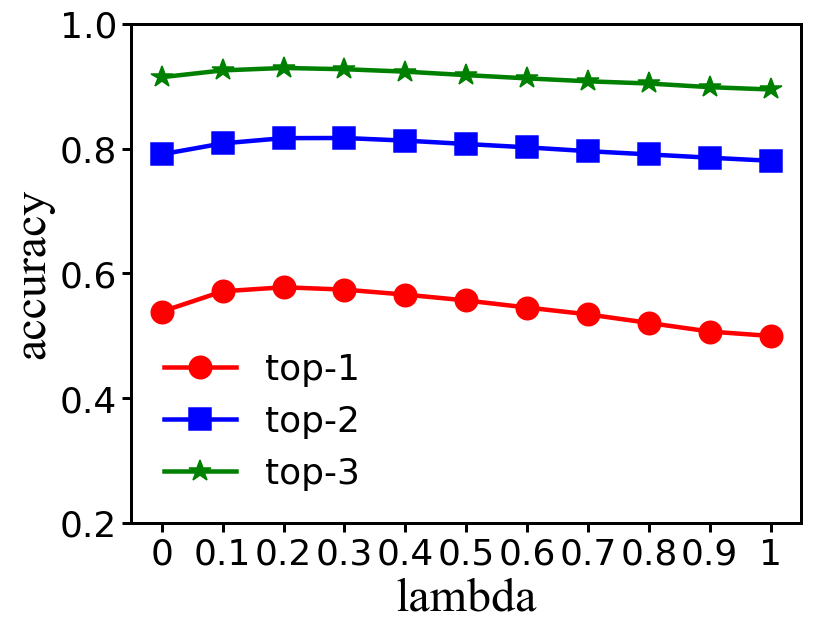}}
\hspace{0.1\textwidth}
\subfigure[Average Precision]{
\includegraphics[width=0.35\textwidth]{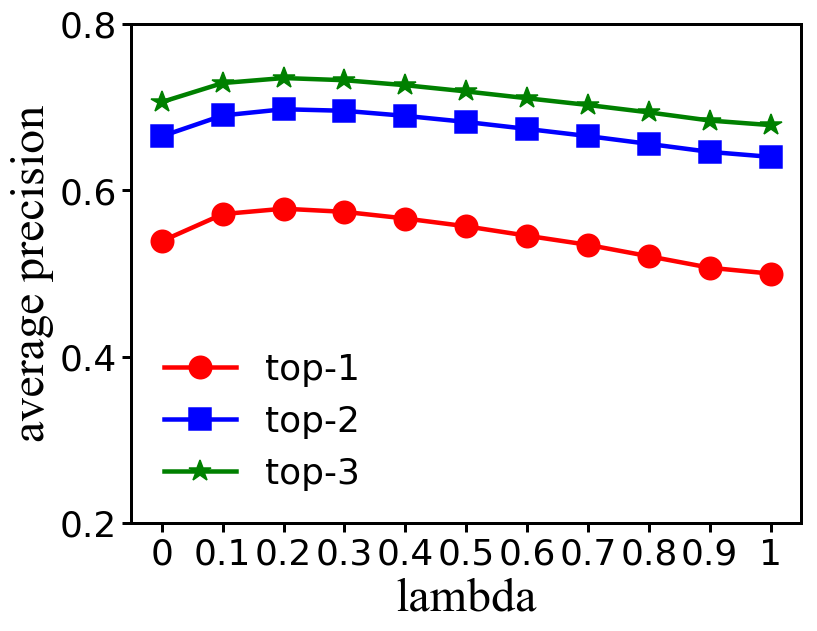}}
\caption{Performance of the joint model with different $\lambda$ on the Taxi data. }
\label{fig:lambda2}
\end{figure*}

\subsection{Comparisons with Baselines}
Our proposed TTDM considers all the passed locations in the query trajectory sequence, and uses the travel time difference to predict next locations. Further, we integrate our model with a Markov model (MM for short) to enhance the prediction performance. Therefore, we evaluate the performance of MM, TTDM, and the joint model (TTDM+MM for short) on the VPR data and the Taxi data, respectively. We predict top-5 next locations and show the results in Tab.~\ref{tab:baseline1} and Tab.~\ref{tab:baseline2}. The best results are highlighted in boldface.

\begin{enumerate}
\item The Markov model performs the worst on both the VPR data and the Taxi data in terms of accuracy and average precision. For example, the top-1 accuracy is 0.417 on the VPR data and 0.5 on the Taxi data. The reason may be that it only models the just passed location and computes the sequential transition probability, without considering the whole prefix locations in the query trajectory sequence.
\item TTDM models all the passed locations in the query trajectory sequence, and considers the travel time from the passed locations to the candidate next location, in which it could capture the effect of the long distance prefix locations. Consequently, TTDM performs better than MM based on accuracy and average precision on both datasets. For example, compared with MM, the top-1 accuracy improves by 26.4\% on the VPR data and by 7.6\% on the Taxi data.
\item MM learns the local sequential transition for a candidate next location and TTDM models the global travel time of all the passed locations to a candidate next location, and the two models contribute differently into the probability of visiting a candidate next location. Hence, the joint model (TTDM+MM) performs the best by incorporating both the local and the global information. For example, compared with MM, the top-1 accuracy improves by 40\% on the VPR data and by 15.6\% on the Taxi data. The top-5 accuracy could reach 0.965 on the VPR data and 0.992 on the Taxi data.
\end{enumerate}

\subsection{Parameter Settings and Tuning}
We set one parameter ($\lambda$) in the joint model, which balances the Markov model and the Travel Time Difference Model. We vary $\lambda$ from 0 to 1 with a step of 0.1, and evaluate the performance of the joint model (TTDM+MM) on the VPR data and the Taxi data. We demonstrate the top-3 accuracy and average precision in Fig.~\ref{fig:lambda1} and Fig.~\ref{fig:lambda2}. As shown in Fig.~\ref{fig:lambda1} (a), the accuracies improve when we increase $\lambda$ from 0 to 0.3, and then start to decrease when we further increase it. The average precisions have the similar trend according to the results in Fig.~\ref{fig:lambda1} (b). Furthermore, the accuracies and average precisions with different $\lambda$ on the Taxi data are shown in Fig.~\ref{fig:lambda2}. The best performances could be obtained when $\lambda$ is equal to 0.2. Note that, the joint model is the same as the TTDM when $\lambda$ is 0, and the same as the MM when $\lambda$ is 1.

\section{Conclusions and Future Work}
\label{conclusion}
In this paper, we have proposed a Travel Time Difference Model (TTDM for short) to predict next locations. TTDM models all the locations in the query trajectory sequence. Specifically, it considers both the shortest travel time and the actual travel time from all the passed locations in the query trajectory sequence to a candidate next location, and leverages the travel  time difference to make next location prediction. Further, we introduce a Markov model which mainly learns the local sequential transitions, and integrate the TTDM with the Markov model to yield a joint model to enhance the prediction performance. We use two real datasets that have different features to evaluate our models via next location prediction. The experiments results show that our TTDM significantly outperforms the Markov model, and the joint model performs the best.

In the future, we aim to integrate our TTDM with more location prediction models, e.g., the mobility pattern embedding model \cite{chen2018mpe}, the recurrent neural network based model \cite{liu2016predicting}. Further, we plan to apply the joint model on more types of traffic data (e.g., Uber ride data, and Didi trajectory data).

\bibliographystyle{IEEEtran}
\bibliography{wenxian}

\end{document}